\documentclass{llncs}
\usepackage{makeidx}  % allows for indexgeneration
\usepackage{algorithm}
\usepackage{algorithmic}
\usepackage{amsmath}
\usepackage{subfigure}
\usepackage{graphicx}
\usepackage{xcolor,colortbl}

\DeclareMathOperator*{\argmax}{arg\,max}
\begin{document}
\frontmatter          % for the preliminaries
\pagestyle{headings}  % switches on printing of running heads
\mainmatter              % start of the contributions
\title{Uniresolution  representations of white-matter data from CoCoMac }
\titlerunning{White Matter network}  % abbreviated title (for running head)
\author{Raghavendra Singh}
\authorrunning{R. Singh} % abbreviated author list (for running head)
\institute{IBM Research India, India\\
ISID Campus, Vasant Kunj, Delhi\\
\email{raghavsi@in.ibm.com}}

\maketitle              % typeset the title of the contribution

\begin{abstract}
Tracing data as collated by CoCoMac, a seminal neuroinformatics database, is at multiple resolutions -- white matter tracts were studied  for areas and their subdivisions by different reports. Network theoretic analysis of this multi-resolution data often assumes that the data at various resolutions is equivalent, which may not be correct. In this paper we propose three methods to resolve the multi-resolution issue such that the resultant networks have connectivity data at only one resolution. The different resultant networks are compared in terms of their network analysis metrics and degree distributions.
\end{abstract}
\section{Introduction}
Tracing studies have been the conventional mechanism for studying white matter tracts in the primate brains. There have been many studies over the past century, and neuroinformatics databases such as  Collation of Connectivity data on the Macaque (CoCoMac)~\cite{cocomac} have made a valiant effort to collate these studies. These collations, and networks extracted from the collations, have been very useful in network theoretic analysis of white matter data, providing a number of remarkable insights into the functioning of the brain including distributed and hierarchical structure of cortex \cite{vanEssen:91}, topological organization of the cortex \cite{Young92}, %indeterminacy of unique hierarchy \cite{hilgetag:96}, 
functional small-world characteristics, optimal set analysis, and multidimensional scaling \cite{compcocomac}, 
small-world characteristics \cite{sporns:zwi},  nonoptimal component placement for wire-length \cite{kaiser:hilgetag}, structural and functional motifs \cite{sporns:kotter}, hub identification and classification \cite{Honey2}, exponential degree distribution, 
%prefrontal areas that are disproportionately  topological central, 
and tightly integrated core subnetwork~\cite{ModSin10}.

A core issue with CoCoMac connectivity data is that it is at various resolutions. For example while Yeterian and Pandya have studied white matter tracts of area {\sf V4}~\cite{YP95},  Lewis and Van Essen have studied tracts of subdivisions of {\sf V4}~\cite{LV00b}; while Walker~\cite{W40} studied area {\sf 24} of the cingulate cortex and its tracts to prefrontal cortex, Vogt et.al.~\cite{VPR87} have studied subareas of {\sf 24}; similar examples abound in CoCoMac data. Analysis which rely on this data either use subsets of the data, for example~\cite{Honey2}, or in the case of more comprehensive dataset~\cite{ModSin10}, incorporate all resolutions of the data.  

Studying multiple resolutions of a data raises the concern that are  all the resolution equivalent? For example with the connectivity network in~\cite{ModSin10}, there is an accompanying hierarchy that codifies the resolutions -- areas are organised to represent the recursive subdivisions. The connectivity data is not only reported for all leaf vertices but also for some internal vertices of the hierarchy. So for this dataset would a connection at an internal vertex be equivalent to a connection at its descendant leaf vertex? If not would the network theoretic analysis results change?

In this paper we are proposing methods to resolve the resolution issue in the network of~\cite{ModSin10}, such that the resultant networks have connectivity data at a single resolution. In other words realise a ``uni''resolution network that has connectivity only at the leaf vertices, not at the internal vertices of a hierarchy. The central problem in realising such networks is  to assign the connectivity data of an internal vertex to which one (or more) of its descendant leaf vertices? 

A simple principle would be of inheritance, that is to assign the connectivity data of internal vertex to each of its descendant leaves. This  creates a network with many edges,  and with the highest resolution possible. The caveat is that  some of the connectivity assigned to the leaves may not exist if traced in the Macaque brain. On the other hand the  principle of disinheritance could also be applied, internal vertices that have connectivity acquire  connectivity data from their descendant vertices. This will create a network with few edges and vertices, but all edges could be traced in primate brain if studied at the right resolution. 

A third more interesting principle is based on Kron reduction~\cite{kron} to remove the internal vertices and reassign its connectivity to leaf vertices. This is followed by sparsification based on electrical resistance~\cite{SpiSri}.  Intuitively Kron reduction preserves paths in the graph, while removing some of the vertices. It has a number of interesting properties, including preserving resistance distance~\cite{kron}. Resistance distance across and edge is the potential distance induced across it when a current is injected at one end of the edge and extracted at the other end. Spielman and Srivastava have proposed a sparsification algorithm that selects edges based on their resistances and produces high quality spectral sparsifiers of weighted graphs~\cite{SpiSri}. In this paper we use Kron and sampling to select for each connection  of an internal vertex, the leaf vertex that inherits it. Though the resulting connectivity may not be traced in Macaque brain, this principle realises networks that are high resolution and sparse. We compare these three principles in terms of known network theoretic metrics. 

\section{Methods and Dataset}
In this section we outline the algorithms and discuss the dataset used. Let us start by establishing the notation, a network is represented as $G=[V,E,W]$, which is a set of vertices $V= \{v\}$, a set of edges $E=\{e_{uv}\}$ where $e_{uv}$ is a directed edge from vertices $u$ to $v$, and a set of weights associated with the edges $W = \{w_e\}$. For the case of unweighted networks $w_e = 1 \; \forall e \in E$, or in short notation $W = {\bf{1}}$. To distinguish between  networks we  use the superscript notation, and capital letters to denote sets unless otherwise stated. $E_{v:}$ represent all edges from $v$, and $E_{:v}$ represent all edges to $v$. $p_{uv}$ represents the shortest path between $u$ and $v$ in $G$. $E_{S,T}$  represents all the edges between the vertices in set $S$ and set $T$. 

An input to our algorithms is the connectivity network $G=[V^G, E^G, W^G]$ where the vertices could denote the brain areas as in~\cite{ModSin10}, or could represent people in an organization. An edge $e^G_{uv}$ could imply the presence of a white matter pathway between brain areas as in~\cite{ModSin10}, or a communication channel between people in an organization. The weights would imply the relative strength of the pathway or of the communication channel\footnote{the network in~\cite{ModSin10} is unweighted}. The other required input  is a hierarchy $T = [V^G,E^T,{\bf{1}}]$. This is a tree graph with no cycles, and a root vertex. The hierarchy could represent the subdivisions of the brain into structurally and functionally consistent brain areas as in~\cite{ModSin10}, or could be the org chart in an organization. Thus the inputs of the algorithms are a network $G$ that has edges at internal vertices of a hierarchy $T$, while  the outputs  are a network $R =[V^R, E^R, W^R]$ and a hierarchy $M=[V^R, E^M,{\bf{1}}]$ such that $R$ does not have edges at the internal vertices of $M$.

The algorithm based on the principle of inheritance is outlined in Algo.\ref{alg1}. The algorithm finds all internal vertices of $G$ that have connectivity, $I$ (in the algorithm), for each of these vertices it finds descendant leaf nodes, $L$, and transfers its connectivity to each of the leaf nodes. The output network $R$ has the same number of vertices as $G$, and a substantially higher number of edges. Output hierarchy $M$ is the same as input hierarchy $T$. \begin{algorithm}
\caption{($R,M$) = Inherit($G,T$)}
\label{alg1}
\begin{algorithmic}
\STATE $R = G$;
\STATE $M = T$
\STATE $I = \{v : v \in V^G \land (\exists \; e^G_{v:} \lor \; \exists \; e^G_{:v}) \;\land\;e^T_{v:}\}$
\FOR{$ u \in I$}
	\STATE $L = \{v \in V^G : \exists\; p^T_{uv} \land \; \not\exists\; e^T(v:)\}$;
	 \FOR{$ v \in L$}
	\STATE $W^R_{v:} = W^R_{v:}+W^G_{u:}$
	\STATE $W^R_{:v} = W^R_{:v}+W^G_{:u}$
	\ENDFOR
	\STATE $W^R_{u:} = 0$;
	\STATE $W^R_{:u} = 0$;
	\STATE $E^R_{u:} = \emptyset$;
	\STATE $E^R_{:u} = \emptyset$;
\ENDFOR
\end{algorithmic}
\end{algorithm}

In Algo.\ref{alg2} the disinheritance principle is used to realise a network with uniresolution connectivity. The algorithm finds all the internal vertices of $G$ that have connectivity $I$, and for each of these vertices it finds descendants leaf nodes $L$, and transfers the connectivity of the leaf nodes to itself. It then removes the leaf nodes from the output network $R$ and the hierarchy $M$.  The output network $R$ has less number of vertices as $G$, and a substantially lower number of edges. 
\begin{algorithm}
\caption{($R,M$) = DisInherit($G,T$)}
\label{alg2}
\begin{algorithmic}
\STATE $R = G$;
\STATE $M = T$G
\STATE $I = \{v : v \in V^G \land (\exists \; e^G_{v:} \lor \; \exists \; e^G_{:v}) \;\land\;e^T_{v:}\}$
\FOR{$ u \in I$}
	\STATE $L = \{v \in V^G : \exists\; p^T_{uv}\; \land \; \not\exists\; e^T_{v:}\; \land \; (\exists \; e^G_{v:} \lor \; \exists \; e^G_{:v}) \}$;
	 \FOR{$ v \in L$}
	\STATE $W^R_{u:} = W^R_{u:}+W^G_{v:}$
	\STATE $W^R_{:u} = W^R_{:u}+W^G_{:v}$
	\STATE $W^G_{u:} = 0; \; \; W^G_{:u} = 0$;
	\STATE $E^R_{u:} = \emptyset; \;\; E^R_{:u} = \emptyset$;
	\STATE $E^G_{u:} = \emptyset;\;\;E^G_{:u} = \emptyset$;
	\ENDFOR
\ENDFOR
\FOR{$u \in V$}
		\STATE $C = \{v \in V^R : \exists\; p^T_{uv}\ \}$;
	 \FOR{$ v \in C$}
	 \STATE $V^R = V^R \setminus v;\;\; V^m = V^m \setminus v$;
	 \ENDFOR	
	 \STATE $E^m_{u:}= \emptyset$
	 \ENDFOR
\end{algorithmic}
\end{algorithm}

The above two algorithms are simple and in either case follow a take-all principle -- either all the descendants inherit equally from their ancestors, or an ancestor disinherits all its descendants. The resultant networks are clearly the opposite end of a spectrum, and there are many networks in the entire range of the spectrum. In the last algorithm we propose that each edge of the original network should be represented by {\em one and only one} edge in the resultant network. Further the network should have the highest resolution possible, in other words the resultant edges should be at the leaf nodes of a hierarchy $M$, with $M==T$. Towards this we propose the KronSampling algorithm in Algo.\ref{alg3}. The intuition behind this algorithm is to find for each edge at an internal vertex in $G$, {\em an} edge at its descendant vertices. To select this winner-take-all edge we use precise ordering of edges based on depth from the root in the hierarchy, and probability based on effective resistance and number of times the edge was reported in the sub-hierarchy. This allows us to realise a network with sparsification guarantees~\cite{SpiSri}.

The algorithm requires three modules, each of which are described next. The Kron reduction of a graph is again a graph whose Laplacian matrix is obtained by the Schur complement of the original Laplacian matrix with respect to a subset of nodes~\cite{kron}. It is defined as follows, let $\mathcal{L}$ be the Laplacian of $G$~\cite{Chu} -- if we consider $W$ as a matrix where the $uv$ element is set to $w_{uv}$ and other elements to zero, and $D$ is a diagonal matrix with elements $d_{uu} = \sum_v w_{uv}$, then the Laplacian matrix is $\mathcal{L} = D-W$. Now let $U$ be a subset of $V$, the Kron reduction of the Laplacian can be the given by,
\begin{equation}
	\mathcal{L^K} := \mathcal{L}_{UU}-\mathcal{L}_{UU^c}\mathcal{L}^{-1}_{U^cU^c}\mathcal{L}_{U^cU} \label{eqn1}
\end{equation}
where $\mathcal{L}_{AB}$ represents the $|A|$ x $|B|$ sub matrix consisting of all entries in $\mathcal{L}$ whose row index is in $A$, and whose column index is in $B$. We can then uniquely associate with Laplacian $\mathcal{L}^K$ a reduced weighted graph $K = [V^K=U, E^K, W^K]$ by letting,
\begin{eqnarray}
W^K_{uv} &=& -\mathcal{L}^K; \; if \; u \ne v \label{eqn2}\\
&=& 0 \;else \label{eqn3}
\end{eqnarray}
The edges in $E^K$ exist wherever $W^K$ is nonzero. $U^c$ represents the compliment of $U$ in $V$. The many properties of Kron reduction are detailed in~\cite{kron} and references therein. The module $K = Kron(G,U)$ takes the graph $G$, finds it Laplacian $\mathcal{L}$, and uses equation(\ref{eqn1}) and then equations(\ref{eqn2}, \ref{eqn3}) to output the reduced network $K$. 

The second module  $R = Resistance(K)$ outputs a network $R=[V^R=V^K,E^R=E^K,W^R]$ whose weights are the effective resistance of the edges of input network $K=[V^K,E^K,W^K]$. If $\mathcal{L}$ is the Laplacian of $K$ then effective resistance is defined by,
\begin{equation}
W^R_{uv} = \mathcal{L}^\dagger_{uu}+ \mathcal{L}^\dagger_{vv}-2 \mathcal{L}^\dagger_{uv}
\end{equation}
where $ \mathcal{L}^\dagger$ is the Moore-Penrose inverse of $ \mathcal{L}$. Again~\cite{kron} and furthermore~\cite{SpiSri} have an excellent discussion on the properties of the effective resistance of a network.

The final module $P = Probability(O,Z)$ takes networks $O$ and $Z$ that have the same vertices, the same edges, but different weights on edges, and outputs a network $P$ with same vertices and edges, and weights defined by,
\begin{equation}
W^P_{uv} = \frac{W^O_{uv}\times W^Z_{uv}}{\sum_{u,v} (W^O_{uv}\times W^Z_{uv})}
\end{equation}
Clearly each weight assigns a probability distribution over the edges in the network $P$.  Note that  if $G$ was unweighted, i.e., $w^g_{uv} =1 \; \forall \;u,v$ as in the network of~\cite{ModSin10}, the weights of the resultant network $R$ for Inherit algorithm will reflect  the number of times connectivity is reported in a sub-hierarchy of $G$, i.e., $w^R_{u,v} = k$ implies that there are $k$ edges between the descendants of $u$ and $v$ in $G$~\footnote{if $G$ is positive weighted this interpretation will still hold, with $w^R_{uv} \propto k$}. Thus $W^R$ can be taken atleast as an indicator of the number of times tracer studies were done for the incident areas, albeit at different resolutions. We will use $W^R$ along with the effective resistance to sample edges.

Having estimated the probability from both the effective resistance and count of times connectivity is reported in a sub-hierarchy, we sort all edges of original network in increasing order of product of their depth first distance from root vertex $Br$. Thus the first edge to be considered is one which is between the farthest leaf vertices of $G$, and all leaf-leaf edges are added to $E^R$. Then among the remaining edges of $G$ the most likely edges are sampled if they do not already exist in $R$ network. 

\begin{algorithm}
\caption{($R,M$) = KronSampling($G,T$)}
\label{alg3}
\begin{algorithmic}
\STATE $R = G$;
\STATE $M = T$;
\STATE $L = \{v \in V^g :  \not\exists\; e^T_{v:}\; \land \; (\exists \; e^G_{v:} \lor \; \exists \; e^G_{:v}) \}$;
\STATE K = Kron(G, L);
\STATE O = Resistance(K);
\STATE Z  = Inherit(G,T);
\STATE P = Probability(O,Z);
\STATE D = DepthFirstSearch(M,1);
\STATE $E^d$ = Sort($E^G$,D);
\STATE $W^R = 0;\; V^R = V^G;\; E^R = \emptyset$;
\FOR{$ e_{uv} \in E^d$}
	\STATE $L_u = \{w \in V^G : \exists\; p^T_{uw}\; \land \; \not\exists\; e^T_{w:}\; \land \; (\exists \; e^G_{w:} \lor \; \exists \; e^G_{:w}) \}$;
	\STATE $L_v = \{w \in V^G : \exists\; p^T_{vw}\; \land \; \not\exists\; e^T_{w:}\; \land \; (\exists \; e^G_{w:} \lor \; \exists \; e^G_{:w}) \}$;
	\STATE $L_u = L_u \cup u$;
	\STATE $L_v = L_v \cup v$;
	\IF{$\not\exists\; e \in E^R_{L_uL_v}$}
		 	\STATE $(s,t)  = \argmax W^p_{L_uL_v}$;
		\STATE $W^R_{st}  = 1$;
		\STATE $E^R = E^R \cup e^R_{st}$;
	\ENDIF
\ENDFOR

\end{algorithmic}
\end{algorithm}
We study the long range network of the Macaque brain as derived by Modha and Singh~\cite{ModSin10}. The network is based on tracing studies of the Macaque brain compiled by the online database CoCoMac~\cite{cocomac}. It covers 383 cortical and sub-cortical brain areas and codes the presence of 6602 directed projections between these areas. The brain areas are arranged in a hierarchal brain map, which is consistent with a recursive parcellation of the brain~\cite{ModSin10}. %In the network studied in~\cite{ModSin10} 32 areas have no connectivity information, rather they are required to hold the hierarchal map together. 
351 of the 383 areas have connectivity; the remaining areas are container or super-areas that hold the hierarchy together.\footnote{We differentiate a super-area from a brain area in that a super-area is sub-divided into brain areas and it does not report any projections.} 
The hierarchal map divides the brain (Br) into  basal ganglia (BG), diencephalon (DiE), and cortex (Cx). Cortex is divided into 6 lobes, temporal (TL\#2), occipital (OC\#2), parietal (Pl\#6), frontal (FL\#2), cingulate gyrus (CgG\#2) and insula (Ins). These super-areas are further sub-divided into other super-areas and brain areas. 
\section{Results and Discussion}
\begin{figure}
\centerline{
\includegraphics[height=3.75in,width=5in]{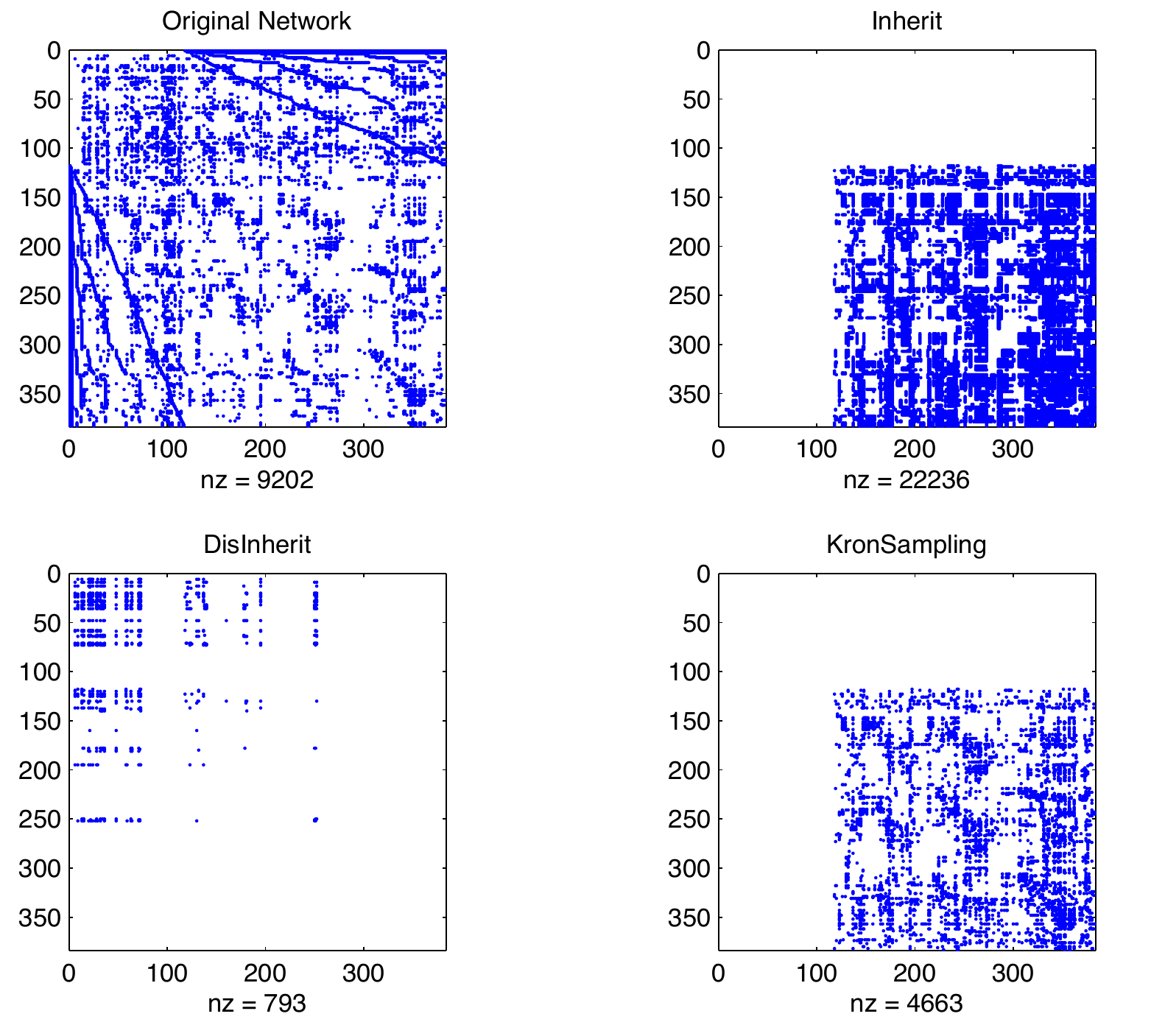} 
 }
 \centerline{
 \includegraphics[height=3.75in,width=5in]{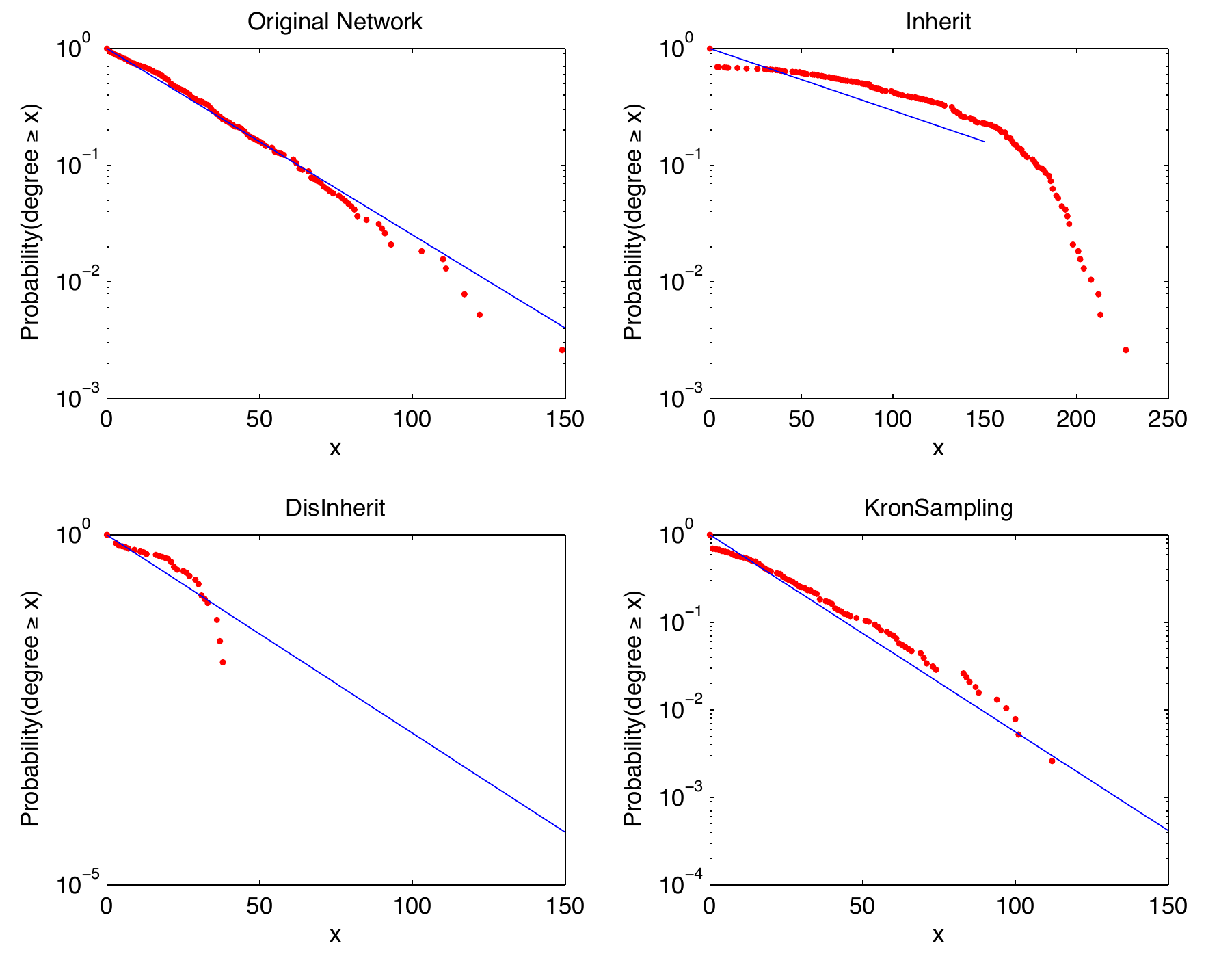}
}
\caption{ The top 4  are spy plots of the adjacency matrix of each network. The matrices have been reordered such that the  internal vertices of the original hierarchy of~\cite{ModSin10} are ordered first, while leaf vertices are ordered last. Thus both Inherit and KronSampling networks have connectivity only between leaf vertices.  Disnherit network, as explained, has connectivity between internal vertices, but the descendants of these vertices can be removed thus again ensuring that connectivity is between leaf vertices of resultant hierarchy.
The bottom 4 plots shows the complementary cumulative distribution of the maximum entropy exponential distribution fit over the range of the data for each network. These plots suggest that the hypothesis of the maximum entropy exponential distribution is consistent with the data particularly for Original and Kron Sampling networks.}
  \label{fg-deg}
\end{figure}

In this section we present the spy plots for all four networks in Fig.~\ref{fg-deg}(top), while (bottom) four plots show the degree distribution of the networks in semi-logy scale fitted to a maximum entropy exponential distribution. In a network, degree of a vertex is the total number of edges that it touches. The tail behavior of the frequency distribution of degrees is a key signature of how connectivity is spread among vertices~\cite{ModSin10}. 
In Tables~\ref{tab4} --\ref{tab7} the top areas for various network theoretic metrics are shown for each network. Besides all other observations we must emphasise that Table~\ref{tab7} shows that the prefrontal cortex predominates the Top-10 table, which is consistent with results in~\cite{ModSin10}. The table was computed using Pajek \cite{pajek}. Due to lack of space we will conclude that we have provided three methods for taking a multi-resolution network and converting it to a single-resolution. In terms of network theoretic analysis we see that the KronSampling network is most similar to the original network.
We must point out that these algorithms are novel and each of them can be used to solve similar multi-resolution problems.

\begin{table}
\begin{center}
\begin{tabular}[t]{|l|l|l|l|l|}
\hline
Metric&Original Network&Inherit&DisInherit&KronSampling\\\hline
\#Verticies&        383&383&66&383\\\hline
\#Edges&       6491&22236&793&4663\\\hline
Density&0.0509&0.3143&0.182&0.0659\\\hline
Reciprocity&   0.4223&0.3567&0.7238&0.4160\\\hline
Diameter&          6&6&4&6\\\hline
Char. path lengt&   2.614&1.932&1.771&2.4836\\\hline
Mean Clustering Coefficient (Directed)& 0.3140&0.5926&0.5066&0.2298\\\hline
\end{tabular}
\caption{Metric for four different networks. The number of vertices reported are the number in the hierarchy, some of them may not have connectivity. Original Network is from~\cite{ModSin10}. The various metrics are also described in the same paper.}
\label{}
\end{center}
\end{table}

\begin{table}[]
%\vspace*{-1in}
\begin{center}
\begin{tabular}{|l|l|r|r|r|r|r|r|r|r|r|r|} \hline
{\em Characteristic} & {\em Rank} $\longrightarrow$ & 1 & 2 & 3 & 4 & 5 & 6 & 7 & 8 & 9 & 10\\ \hline
Integrator & In-Degree & 
\cellcolor{red}32& \cellcolor{red}46& \cellcolor{red}12o& \cellcolor{red}12l& \cellcolor{red}11& \cellcolor{blue}24& \cellcolor{pink}F7& \cellcolor{red}14& \cellcolor{red}8A& \cellcolor{pink}LIP\\
 & In-Closeness & 
\cellcolor{red}46& \cellcolor{red}12o& \cellcolor{red}32,11& \cellcolor{blue}24& \cellcolor{red}12l& \cellcolor{blue}MD& \cellcolor{red}8A& \cellcolor{blue}23c& \cellcolor{red}8B& \cellcolor{pink}LIP,F7\\
& Authorities & 
 \cellcolor{red}32& \cellcolor{red}12o& \cellcolor{red}46& \cellcolor{red}11& \cellcolor{red}12l& \cellcolor{blue}24& \cellcolor{red}14& \cellcolor{pink}F7& \cellcolor{blue}MD& \cellcolor{red}9\\ \hline
Distributor & Out-Degree & 
 \cellcolor{red}46& \cellcolor{blue}24& \cellcolor{green}TF& \cellcolor{red}9& \cellcolor{red}13& \cellcolor{red}13a& \cellcolor{green}TH& TE,LIP& \cellcolor{pink}PGm& \cellcolor{green}V2\\
 & Out-Closeness & 
\cellcolor{red}46& \cellcolor{blue}24& \cellcolor{green}TF& \cellcolor{green}TE& \cellcolor{red}9& \cellcolor{green}TH& \cellcolor{pink}LIP& \cellcolor{pink}PGm& 23,PM\#3,45& \cellcolor{red}12\\
& Hubs & 
\cellcolor{red}46& \cellcolor{blue}24& \cellcolor{red}9& \cellcolor{green}TF& \cellcolor{green}TE& \cellcolor{green}TH& \cellcolor{red}13& \cellcolor{red}32& \cellcolor{blue}23& \cellcolor{blue}PM\#3\\ \hline
Intermediary & Betweenness & 
\cellcolor{blue}24& \cellcolor{red}46& \cellcolor{pink}LIP& \cellcolor{red}13a& \cellcolor{blue}MD& \cellcolor{red}32& \cellcolor{green}TF& \cellcolor{green}PIT& \cellcolor{red}13& \cellcolor{red}PS\\ 
& PageRank & 
\cellcolor{red}32& \cellcolor{blue}MD& \cellcolor{red}46& \cellcolor{green}36r& \cellcolor{green}PIT& \cellcolor{red}12o& \cellcolor{blue}24& \cellcolor{blue}23c& \cellcolor{red}12l& \cellcolor{red}11\\ \hline
\end{tabular}

\end{center}
\caption{Top ten brain areas according to several metrics of topological centrality for the directed version of the Original network~\cite{ModSin10}. The cells are color coded according to the following scheme. If the area is a sub-area of pre frontal cortex (PfC) in~\cite{ModSin10} then its coloured red, else if sub-area of motor, parietal lobe, or Insular Cortex (6\#1, M1,  Pl\#6  or Insula) its coloured pink. If area is a sub-area of temporal or occipital lobe (TL\#2 or OC\#2) then its coloured green, and remaining limbic areas (CgG\#2, Tha, BG) are coloured blue. Cells that have multiple areas are not coloured.
}\label{tab4}
\end{table}

\begin{table}[]
%\vspace*{-1in}
\begin{center}
\begin{tabular}{|l|l|r|r|r|r|r|r|r|r|r|r|} \hline
{\em Characteristic} & {\em Rank} $\longrightarrow$ & 1 & 2 & 3 & 4 & 5 & 6 & 7 & 8 & 9 & 10\\ \hline
Integrator & In-Degree & 
 \cellcolor{pink}F7& \cellcolor{blue}24c& \cellcolor{blue}24b& \cellcolor{blue}24d& \cellcolor{blue}24a& \cellcolor{green}TPag& \cellcolor{red}11m& \cellcolor{green}TPg& \cellcolor{red}12o& \cellcolor{red}46v\\
 & In-Closeness & 
\cellcolor{green}TPdgv& \cellcolor{green}TPdgd& \cellcolor{green}TPag& \cellcolor{green}TPg& \cellcolor{green}36p& \cellcolor{red}12o& \cellcolor{red}12l& \cellcolor{red}11l& \cellcolor{red}11m& \cellcolor{red}46v\\
& Authorities & 
\cellcolor{blue}24b& \cellcolor{blue}24c& \cellcolor{blue}24a& \cellcolor{blue}24d& \cellcolor{green}TPag& \cellcolor{red}12o& \cellcolor{green}TFL& \cellcolor{green}TPg& \cellcolor{green}TFM& \cellcolor{pink}F7\\ \hline
Distributor & Out-Degree & 
\cellcolor{blue}24c& \cellcolor{blue}24b& \cellcolor{blue}24d& \cellcolor{blue}24a& \cellcolor{pink}PFG\#1& \cellcolor{pink}PF\#1& \cellcolor{pink}Idg& \cellcolor{pink}Iam& \cellcolor{red}PS& \cellcolor{pink}Ial\\
 & Out-Closeness & 
\cellcolor{pink}PFG\#1& \cellcolor{pink}PF\#1& \cellcolor{blue}24c& \cellcolor{blue}24d& \cellcolor{blue}24b& \cellcolor{blue}24a& \cellcolor{green}TPdgv& \cellcolor{green}TPdgd& \cellcolor{green}TPag& \cellcolor{green}TPg\\
& Hubs & 
\cellcolor{blue}24c& \cellcolor{blue}24b& \cellcolor{blue}24d& \cellcolor{blue}24a& \cellcolor{red}PS& \cellcolor{green}CITv& \cellcolor{pink}Iam& \cellcolor{pink}Idg& \cellcolor{red}45A& \cellcolor{red}45B\\ \hline
Intermediary & Betweenness & 
\cellcolor{blue}24c& \cellcolor{blue}24b& \cellcolor{blue}24d& \cellcolor{blue}24a& \cellcolor{pink}PF\#1& \cellcolor{red}13a& \cellcolor{pink}PFG\#1& \cellcolor{red}PS& \cellcolor{pink}F7& \cellcolor{pink}F2\\
& PageRank & 
\cellcolor{blue}MDpm& \cellcolor{blue}MDfi& \cellcolor{blue}24c& \cellcolor{blue}24b& \cellcolor{blue}MDpc& \cellcolor{blue}24d& \cellcolor{blue}MDmf& \cellcolor{blue}MDdc& \cellcolor{blue}MDcd& \cellcolor{blue}24a\\ \hline
\end{tabular}

\end{center}
\caption{Top ten brain areas according to several metrics of topological centrality for the directed version of Inherit network. Colour coding as in Table~\ref{tab4}.}\label{tab5}
\end{table}
\begin{table}[]
%\vspace*{-1in}
\begin{center}
\begin{tabular}{|l|l|r|r|r|r|r|r|r|r|r|r|} \hline
{\em Characteristic} & {\em Rank} $\longrightarrow$ & 1 & 2 & 3 & 4 & 5 & 6 & 7 & 8 & 9 & 10\\ \hline
Integrator & In-Degree & 
  \cellcolor{red}FD\#1& \cellcolor{red}8& \cellcolor{pink}6\#1& \cellcolor{pink}7\#1& \cellcolor{blue}Tha& \cellcolor{blue}23& \cellcolor{blue}24& \cellcolor{green}STS& \cellcolor{red}PFCorb& \cellcolor{green}Per\#1\\
 & In-Closeness & 
\cellcolor{red}FD\#1& \cellcolor{blue}Tha& \cellcolor{red}8& \cellcolor{pink}6\#1& \cellcolor{pink}7\#1& \cellcolor{blue}23& \cellcolor{blue}24& \cellcolor{green}STS& \cellcolor{red}PFCorb& \cellcolor{green}Per\#1\\
& Authorities & \#
\cellcolor{red}FD\#1& \cellcolor{red}8& \cellcolor{pink}7\#1& \cellcolor{blue}23& \cellcolor{blue}Tha& \cellcolor{pink}6\#1& \cellcolor{green}PHC& \cellcolor{blue}24& \cellcolor{green}STS& \cellcolor{green}Per\#1\\ \hline
Distributor & Out-Degree & 
  \cellcolor{pink}6\#1& \cellcolor{red}FD\#1& \cellcolor{pink}7\#1& \cellcolor{pink}Insula& \cellcolor{red}8& \cellcolor{blue}24& \cellcolor{green}OA& \cellcolor{green}STS& \cellcolor{green}TE& \cellcolor{blue}Tha\\
 & Out-Closeness & 
\cellcolor{pink}6\#1& \cellcolor{red}FD\#1& \cellcolor{pink}7\#1& \cellcolor{pink}Insula& \cellcolor{red}8& \cellcolor{green}OA& \cellcolor{green}STS& \cellcolor{blue}24& \cellcolor{green}TE& \cellcolor{blue}Tha \\
& Hubs & 
\cellcolor{red}FD\#1& \cellcolor{pink}7\#1& \cellcolor{pink}Insula& \cellcolor{pink}6\#1& \cellcolor{blue}24& \cellcolor{green}STS& \cellcolor{red}8& \cellcolor{blue}Tha& \cellcolor{red}PFCorb& \cellcolor{green}TE\\ \hline
Intermediary & Betweenness & 
 \cellcolor{pink}6\#1& \cellcolor{blue}Tha& \cellcolor{red}8& \cellcolor{red}FD\#1& \cellcolor{green}OA& \cellcolor{green}TE& \cellcolor{pink}7\#1& \cellcolor{blue}Amyg& \cellcolor{green}V2& \cellcolor{green}STS\\
& PageRank & 
\cellcolor{blue}Tha& \cellcolor{pink}6\#1& \cellcolor{red}8& \cellcolor{blue}Cd& \cellcolor{red}FD\#1& \cellcolor{blue}23& \cellcolor{pink}7\#1& \cellcolor{green}TE& \cellcolor{green}STS& \cellcolor{blue}Amyg\\ \hline
\end{tabular}

\end{center}
\caption{Top ten brain areas according to several metrics of topological centrality for the directed version of DisInherit network.Colour coding as in Table~\ref{tab4}.
}\label{tab6}
\end{table}
\begin{table}[]
%\vspace*{-1in}
\begin{center}
\begin{tabular}{|l|l|r|r|r|r|r|r|r|r|r|r|} \hline
{\em Characteristic} & {\em Rank} $\longrightarrow$ & 1 & 2 & 3 & 4 & 5 & 6 & 7 & 8 & 9 & 10\\ \hline
Integrator & In-Degree &  \cellcolor{red}14r& \cellcolor{red}32& \cellcolor{red}11m& \cellcolor{red}12o& \cellcolor{red}12l& \cellcolor{red}13a& \cellcolor{pink}PF\#1& \cellcolor{red}10o& \cellcolor{green}TFL& \cellcolor{red}8Ac\\ 
 & In-Closeness & 
 \cellcolor{red}12l& \cellcolor{red}14r& \cellcolor{red}13a& \cellcolor{red}10o& \cellcolor{red}12o& \cellcolor{red}11m& \cellcolor{red}46v& \cellcolor{pink}PF\#1& \cellcolor{green}36r& \cellcolor{green}PITv\\ 
 & Authorities & 
  \cellcolor{red}12o& \cellcolor{red}12l& \cellcolor{red}32& \cellcolor{red}14r& \cellcolor{red}11m& \cellcolor{red}13a& \cellcolor{green}TFL& \cellcolor{red}10o& \cellcolor{green}36r& \cellcolor{blue}23c \\ \hline
Distributor & Out-Degree & 
 \cellcolor{red}13a& \cellcolor{blue}24a& \cellcolor{green}TFL& \cellcolor{red}46dr& \cellcolor{red}12l& \cellcolor{red}D9& \cellcolor{green}TH& \cellcolor{pink}Iai& \cellcolor{red}11m& \cellcolor{pink}F5\\ 
 & Out-Closeness & 
\cellcolor{red}45B& \cellcolor{red}13a& \cellcolor{red}12l& \cellcolor{red}D9& \cellcolor{red}46dr& \cellcolor{pink}PGm& \cellcolor{green}TFL& \cellcolor{blue}24a& \cellcolor{green}TH& \cellcolor{green}TPg\\
& Hubs & 
\cellcolor{red}13a& \cellcolor{red}12l& \cellcolor{green}TFL& \cellcolor{red}D9& \cellcolor{blue}24a& \cellcolor{red}11m& \cellcolor{red}14r& \cellcolor{red}10o& \cellcolor{pink}Iai& \cellcolor{green}TH\\ \hline
Intermediary & Betweenness & 
\cellcolor{red}13a& \cellcolor{blue}24a& \cellcolor{pink}PF\#1& \cellcolor{green}TFL& \cellcolor{green}PITv& \cellcolor{red}12l& \cellcolor{pink}LIPi& \cellcolor{red}8Ac& \cellcolor{red}32& \cellcolor{green}V2\\
& PageRank & 
\cellcolor{blue}MDcd& \cellcolor{red}32& \cellcolor{green}PITv& \cellcolor{red}14r& \cellcolor{red}12l& \cellcolor{red}13a& \cellcolor{red}12o& \cellcolor{red}11m& \cellcolor{blue}24a& \cellcolor{blue}23c\\ \hline
\end{tabular}
\end{center}
\caption{Top ten brain areas according to several metrics of topological centrality for the directed version for KronSampling network.Colour coding as in Table~\ref{tab4}.
}\label{tab7}
\end{table}

{\small
\bibliographystyle{splncs03}
\bibliography{bih2015}
}
\end{document}